# Using Artificial Intelligence to Support Compliance with the General Data Protection Regulation

John Kingston,

University of Brighton

Watts Building, Brighton BN2 4GJ

j.k.kingston@brighton.ac.uk

01273 642938

## Abstract

The General Data Protection Regulation (GDPR) is a European Union regulation that will replace the existing Data Protection Directive on 25 May 2018. The most significant change is a huge increase in the maximum fine that can be levied for breaches of the regulation. Yet fewer than half of UK companies are fully aware of GDPR – and a number of those who were preparing for it stopped doing so when the Brexit vote was announced.

A last-minute rush to become compliant is therefore expected, and numerous companies are starting to offer advice, checklists and consultancy on how to comply with GDPR. In such an environment, artificial intelligence technologies ought to be able to assist by providing best advice; asking all and only the relevant questions; monitoring activities; and carrying out assessments.

The paper considers four areas of GDPR compliance where rule based technologies and/or machine learning techniques may be relevant:

- Following compliance checklists and codes of conduct;
- Supporting risk assessments;
- Complying with the new regulations regarding technologies that perform automatic profiling;
- Complying with the new regulations concerning recognising and reporting breaches of security.

It concludes that AI technology can support each of these four areas. The requirements that GDPR (or organisations that need to comply with GDPR) state for explanation and justification of reasoning imply that rule-based approaches are likely to be more helpful than machine learning approaches. However, there may be good business reasons to take a different approach in some circumstances.

# 1 Introduction

The General Data Protection Regulation (GDPR) is a European Union regulation that will replace the existing Data Protection Directive 95/46/EC on 25 May 2018. The GDPR is directly applicable in each Member State and will also apply to organisations outside the European Union that process data about EU citizens.

The GDPR introduces a number of changes from the older regulation. The most significant change is a huge increase in the maximum fine that can be levied for breaches of the regulation – 'lesser' breaches have a maximum fine of €10 million or 2% of the organisation's turnover; the upper limit for serious violations is twice as high. As an illustration, TalkTalk were fined £400,000 in 2015 for allowing hackers to access customer data; under GDPR, the likely fine would have been £59 million (Leyden 2017).

Other notable changes include:

- Individuals must give 'clear and affirmative consent' to their data being processed. Ticking a box is adequate; staying silent on the subject is not.
- IP addresses and mobile device IDs are now unquestionably defined as personal data (their status was debatable under the Data Protection Directive).
- Genetic data and biometric data are classified as "sensitive personal data".
- There is a strong incentive to pseudonymise data sets; such data is still classed as personal data but is exempted from various requirements of the GDPR. Encryption is also encouraged.
- The "right to be forgotten" established by the European Court of Justice (Google Spain v. AEPD 2014) has been extended to become a "right of erasure"; it will no longer be sufficient to remove a person's data from search results when requested to do so, data controllers must now erase that data. However, if the data is encrypted, it may be sufficient to destroy the encryption keys rather than go through the prolonged process of ensuring that the data has been fully erased.

The GDPR does relax a couple of requirements that were in the previous Data Protection Directive:
- Controllers will no longer be required to register their processing activities with a Data Protection Authority (DPA) in each member state. Instead, the GDPR imposes strict requirements on controllers to maintain their own detailed records of processing.
- Cross-border transfers of data (to certain countries) can now be conducted according to standard contractual clauses without prior approval by DPAs, or can be based on certifications. However, it is now explicitly illegal to transfer personal data out of the EU in response to a legal requirement from a third country.

These changes require significant efforts from many data processing organisations to comply; yet according to a survey of IT professionals conducted in February 2017 (Ashford 2017), less than half of UK companies are fully aware of GDPR – and a number of those who were preparing for it stopped doing so when the Brexit vote was announced. However, GDPR will become law some time before the earliest date when the UK might leave the European Union; some of its provisions apply to data processors in non-EU states; and the UK Information Commissioner's Office has stated that it will introduce a law similar to GDPR post-Brexit. A last-minute rush to become compliant is therefore expected, and numerous companies are starting to offer advice, checklists and consultancy on how to comply with GDPR.

In such an environment, artificial intelligence technologies ought to be able to assist with compliance by providing best advice; asking all and only the relevant questions; and carrying out assessments. This paper investigates four parts of GDPR where AI technology can simplify compliance; two are primarily intended for use before the regulation comes into force, two while it is in force. These are:

1. Checklist interpretation: automating a compliance checklist and offering guidance on key sections.
2. Risk analysis: identifying activities that are high-risk, medium-risk or low-risk according to the GDPR's criteria and taking compliance measures accordingly
3. Giving adequate replies to requests for information about automatic profiling.
4. Identification and risk analysis of potential or actual data breaches.

## 2 Artificial Intelligence technologies: rule-based and machine learning

Before discussing how Artificial Intelligence technology can be applied to support compliance with GDPR, a brief overview of AI technology is needed. There are various technologies that can be classified as AI technologies. Many of them can be loosely classified into two categories: *rule-based* technologies and *machine learning* technologies.

Rule-based technologies include 'expert systems' based on forward and/or backward chaining 'production rules' and/or object-oriented programming. Their key characteristics are:

- A source of rules is required. This is usually one or more experienced humans, or one or more documents describing policy or regulations.
- The system must be designed following knowledge engineering principles, and the rules must be programmed into the system.
- The AI system then asks users questions and reasons with the answers in order to make decisions. A decision tree or similar structure is often used; if it is, the AI system will ask users all and only the relevant questions, omitting any questions which are unnecessary because of previous answers.
- The AI system can easily give explanations of its reasoning if required, by describing the rules that were triggered; the information that triggered them; and their conclusions.

For example, the X-MATE system (Kingston 1991) applies rule-based technology to mortgage application assessment. Individual rules identify combinations of information that increase the risk of mortgage default (or rather, increase one of three risks: that the applicant is poor at budgeting; that the applicant's disposable income may drop significantly due to changing circumstances; or deliberate fraud is being attempted). Each rule concludes whether a low, medium, high or very risk has been identified. These risks are then combined (by simple addition) to produce an overall risk score. On a test set of 50 cases obtained from the Bad Debts department of a top ten UK building society, with the risk threshold set just high enough for zero false positives, 95% of defaulters were identified by this system.

More recent examples of the commercial application of rule based technologies include identifying procurement fraud (Dhurandar et al 2015) and monitoring large numbers of gas turbines, steam turbines, and generators (Thompson et al 2015).

Machine learning technologies include neural networks, data mining, and genetic algorithms. Their key characteristics are:

- A large-scale source of past data is required. Some of these data are used as the training set from which associations are learned, others for validation.

- Once some parameters have been set correctly for optimal learning, the system learns associations by itself; no further programming is required.
- The associations may be presented to the user as individual links or as clusters.
- The AI system cannot explain its reasoning beyond giving information about statistical correlation of data items.

An example of machine learning is an analysis conducted of decisions made by the European Court of Human Rights which attempted to predict the outcome of cases by applying machine learning techniques to the textual content and the components of cases (Aletras et al 2016). The researchers produced a system that was 79% accurate in predicting case outcomes. They discovered that the formal facts of the case were the strongest predictor of outcomes (as would be expected); however, the topics discussed also had some predictive power. For example, when discussing whether Article 3 ("No one shall be subjected to torture or to inhuman or degrading treatment or punishment") had been violated, the presence in the case of words relating to (positive) state obligations (e.g. 'protection', 'quashed' or 'compensation'); or words relating to detention conditions (including 'access' and 'overcrowding') or to treatment by state officials ('subjected', 'force') were all indicative of a violation being found, while three further sets of words indicated that a violation was unlikely to be found.

Machine learning is both powerful and inherently often inaccurate because it makes use of all the associations that exist in the training set, whether or not they are explicit and whether or not they are well-justified. Continuing the above example, three of the 'topic' words that were found to have predictive value were the names of countries. It may be that ECtHR cases involving certain countries are statistically more or less likely to result in particular decisions, but it is highly questionable whether this should be used as a predictive factor for future cases.

## 3 Preparing for GDPR: an AI-supported checklist

Various organisations including the Information Commissioner's Office (2017) have published text-based checklists for organisations to use when preparing to comply with GDPR. The one that will be discussed in this section (Evans (2016)) was produced by a firm of solicitors, and includes the actual text of the Regulation alongside the steps that organisations are advised to take.

One of the items on the checklist concerns data protection training. The checklist says:

"Organisations should implement a training programme covering data protection generally and the areas that are specifically relevant to their organisations, and implement a policy for determining when training should take place and when refresher training should be carried out and a process for recording when training has been completed"

The relevant section of the GDPR is Article 39:

"Data Protection Officers are under a specific obligation to implement appropriate training. Although not an express obligation for organisations where DPOs are not required, we consider it to be almost impossible to demonstrate that an organisation is able to achieve compliance without policies setting out how to comply coupled with training to bring those policies to life."

How can AI technology help organisations to implement this point from the checklist?

1. Assessment. Organisations need to determine what training is appropriate; when training should take place; and when refresher training should be carried out. Each of these is an assessment task, and AI technologies are known to be good at assessment tasks (both examples in Section 1 are assessing how likely it is that individual cases fall into certain categories).

2. Asking all and only the relevant questions. Government legislation is often made more complex by cross-referencing i.e. in order to make a decision about one clause of the regulation, it's necessary to decide about another clause, which may refer to a third clause, and so on. Dealing with cross-references within an information or knowledge system is a long-standing problem – Routen (1989) argues that any knowledge representation that fails to deal with such links will inevitably be impoverished – yet it is vital that any decision support system knows when to stop asking questions as well as which questions must be asked.

A good way to deal with this issue is to ask question in reverse order of dependency: rather than starting by trying to decide whether clause A applies, where A refers to (and depends on) B and B depends on C, the system starts by asking whether C applies. If C does not apply, the system never asks questions about B or A, thus saving the user time and effort. In this example, organisations need to decide whether they are required to designate a Data Protection Officer (DPO). This is covered by a different part of the checklist; the underlying parts of the Regulation are Articles 37-39 and Recital 97. So an AI program that supported decisions about complying with the training clause would start by asking questions to determine whether organisations needed to comply with the requirement to designate a DPO. If they do not, then the questions about training will never be asked.

The knowledge that informs decision making is likely to come from the organisation's existing senior staff; from the organisation's existing policies; or from a consultant, coupled with the wording of GDPR and any other relevant regulations. The human/document knowledge sources and the requirement to ask all and only the relevant questions suggest that the most appropriate AI technology will be rule-based.

Who will be the users of such an AI system? It is likely that each organisation will carry out risk assessments once only, with the results being documented in training material or in a policy. This makes the development of an AI decision support system unnecessarily complex for a single organisation; however, it would be a useful tool for a consultancy organisation that helps many organisations make such decisions.

### 3.1 AI-supported checklist: compliance patterns

Having decided on a rule-based approach, the question of knowledge representation arises: how should the compliance actions taken to address the provisions of GDPR be represented within the rules? A wide range of options is available, from simply copying the questions from the checklist to developing a full argumentation-based framework that justifies the solutions that the system is offering. Methodologies for developing legal knowledge based systems based on various representations are also available, such as Nwana et al (1991) which focuses on the knowledge acquisition process; van Engers (2006) which focussed on the drafting of new legislation; van Kralingen et al (1999); and Al-Abdulkarim (2016).

In practice, organisations who are guided by the system may want to see little more than the checklist questions. However, some questions will trigger discussions about how much (or how little) organisations need to do to comply with the GDPR, or how well the system's suggested approach meets the GDPR's requirements. Because of this, it is sensible for the system to employ a detailed underlying knowledge representation that the consultants who operate the system can refer to if needed.

Most modern "rule-based" software applications in the legal domain use an argumentation-based framework. Examples can be seen in Al-Abdulkarim et al (2016) and Gordon (2013). The suggested formalism below is based on the "compliance patterns" proposed by Muthuri et al (2017).

The "compliance patterns" approach involves detailed representation of facts and arguments at six stages: domain classification; confrontation; opening; legal interpretation argumentation; and closing. The example below described the 'opening' stage for the 'training' question outlined above:

1. Legal claim. The legal claim being made by the supervisory authority is that the organisation has not done enough to comply with Article 39 of the GDPR:
    a. General rule premise: Compliance with EU Regulations is mandatory for organisations that conduct business within one or more states of the European Union.
    b. Performance premise: To perform the activity in question, DPOs must subscribe to policies on how to comply with GDPR and offer training to bring those policies to life.
    c. Warrant: The organisation is in breach of the GDPR if their DPO does not offer such training.
    d. Conclusion: Therefore training must be implemented and offered.
    e. Else: the organisation is in breach of Article 39.
2. Legal action:
    a. Established rule premise: Where a valid legal claim exists under GDPR, supervisory authorities have a right to take action [Articles 51-59, especially Article 58].
    b. Remedies premise: the organisation is potentially liable to comply with legitimate instructions from the supervisory authority and to pay any fines levied pursuant to Article 83.
    c. Violation premise: the organisation violates the GDPR by not complying with its requirements, or not doing so adequately.
    d. Conclusion: therefore the organisation is liable to pay/perform remedies decided by the supervisory authority.

Now let us assume that the organisation wants to challenge the legal claim and/or action above on the grounds that it isn't required to appoint a DPO by GDPR. This can be represented as follows:

3. Exceptional case:
    a. Exception premise: if the organisation is not a public authority; does not have "regular and systematic monitoring of data subjects on a large scale" as one of its core activities; and does not conduct large-scale processing of "special categories of personal data"
       Then the organisation is not required to appoint a Data Protection Officer. [Article 37]
    b. The case cited is an exception
    c. Conclusion: Therefore the organisation is exempted from the Article 39 requirement for the DPO to offer training.

This example illustrates many of the benefits of such a knowledge representation within an AI-based compliance support system:

- The representation makes it easy to locate articles in GDPR that underlie certain requirements.
- The representation is fairly detailed but is relatively easy to encode in a rule-based AI system. It could be encoded using a meta-rule architecture: each clause of the Opening could be represented as a fact within the system, and meta-rules could reason with these facts to draw conclusions and to generate explanations.

- The representation highlights points on which the GDPR is unclear, and which could therefore be open to interpretation. For example, the Warrant from the Legal Claim asserts that the organisation is in breach of the GDPR if their DPO does not offer suitable training. But what if someone other than the DPO offers the training? Can training be bought in from outside providers?

Even in the above brief analysis, several possible points that require interpretation can be identified:

- DPOs are required to offer training, but do staff have to accept the offer?
- Which staff need to be trained?
- If multiple jurisdictions are involved, which supervisory authority/ies should take action?
- Who or what decides whether training is adequate or suitable?
- If an organisation does not require a DPO, does it truly have no requirement to train its staff in data processing issues? Might there be a requirement under contract or tort law?
- What do words such as "regular" and "large-scale" mean?

Highlighting decisions that are open to interpretation is an essential first step in deciding how much needs to be done to comply with the GDPR. If organisations want to interpret a statement in a way that exempts them from some aspect of compliance, they can carry out a simple test of the robustness of their view by expressing their view as an Exceptional Case; this may highlight weaknesses or issues requiring interpretation in their own arguments.

# 4 Risk assessment

Risk assessment and risk analysis are fundamental underlying concepts of the GDPR. The regulation refers frequently to "high-risk" processing activities; Article 33 places a responsibility on data controllers to conduct an assessment for such activities, and Articles 32 and 34 require additional procedures from a data controller when processing high-risk data.

Risk analysis is also supposed to underlie the measures that data controllers take to comply with the regulation. Article 30 of the GDPR requires controllers to "ensure a level of security appropriate to the risk." Article 22 instructs controllers to "implement appropriate technical and organisational measures" that reflect "the risks […] for the rights and freedoms of individuals." Article 23, which deals with the requirement to implement privacy protection at the design stage, states that the chosen privacy measures should reflect the risk and context of the controller's processing activities, as well as the available technology and cost of implementation. And Article 79 instructs supervisory authorities to set penalties for non-compliance "having regard to technical and organisational measures implemented pursuant to Articles 23 and 30" – so risk assessment also underlies the assessment of penalties under the GDPR. (Maldoff 2017).

A collection of possible "technical and organisational measures" are spelled out in Article 32. They are:

- The pseudonymisation and encryption of personal data.
- The ability to ensure the ongoing confidentiality, integrity, availability and resilience of processing systems and services.
- The ability to restore the availability and access to personal data in a timely manner in the event of a physical or technical incident.
- A process for regularly testing, assessing and evaluating the effectiveness of technical and organizational measures for ensuring the security of the processing.

The above requirements mean that organisations that process data must assess the risk posed to data subjects by their data processing activities and then do one (or both) of two things. They must either select appropriate measures to mitigate those risks, or they must identify a suitable code of conduct/certification and adapt their processing activities to comply with it (Articles 40 and 42). In either case, they must allow sufficient time after making their selection to implement the necessary changes before May 2018.

It is therefore likely that there will soon be a large demand for scarce expertise on data processing risk; the demand may even be comparable to the demand for programming skills to deal with Millennium Bug risks in 1997-1999. In such situations, AI technology can help by encoding expertise from one or a few individuals and making it more widely available; risk assessment is a task to which AI solutions have often been applied, such as the X-MATE example above. As in Section 3 above, risk assessment is a one-time decision and so any AI decision support system is likely to be developed by a consultancy rather than by data processing organisations.

The key knowledge in such a system will be an assessment of the risks, and a link between levels of risk and appropriate mitigation measures. The underlying knowledge representation must be capable of justifying the system's recommendations on both these aspects. Once again, therefore, the preferred implementation technique is likely to be rule-based.

What knowledge representation should be used? The two stages should be considered separately:

- The risk assessment stage considers the risks of data loss to the rights and freedoms of individuals. In principle, therefore, it consists of a mapping between the type and volume of data being processed and the rights and freedoms that would be breached if that data was leaked or lost.

    In practice, this is likely to require an argumentation-based knowledge representation derived from previous legal decisions regarding lost data, describing what the court considered to be the likely consequences of that loss. As in Section 3, there may be circumstances where organisations claim that exceptions arise; these too can be encoded in the same representation.

    The knowledge that is required for such a system may be reasonably extensive, and may change as case law is established on the basis of GDPR. However it is not organisation-dependent, so as in Section 3, it may be that only consultancies who advise multiple organisations on GDPR can make a business case for developing such a system.

- The linking between levels of risk and appropriate mitigation measures should reflect "the available technology and cost of implementation" as well as the processing activities. This, therefore, is both a legal decision and a business decision. If precedents are available stating that a particular mitigation measure is suitable for a particular level of risk, and the company is happy to expend the money and time required to implement that measure, then that would be a good legal justification for applying that measure. If the company argues that certain mitigation measures are beyond its budget, then some method of weighing the need for mitigating measures against budgetary constraints is needed.

    In theory, this task could require significant levels of both legal and business knowledge. In practice, it is likely that a consultant-developed AI system would simplify this decision from a point-by-point selection of mitigating actions to a selection between a small set of standard codes of conduct or certifications. The extra cost to the organisation in implementing one or two mitigation approaches that it might not need will usually be outweighed by the reduced time needed for legal consultancy and the increased certainty that their chosen mitigating actions will meet the requirements of the GDPR.

From a commercial perspective, a consultancy that offers an AI system to support selection of risk-mitigating actions might develop an 80-20 rule-based AI system. There is a knowledge engineering

'rule of thumb' (based on the Pareto Principle, originally observed by Pareto (1897) and first applied to quality control by Juran (1951)) that developing a system that covers 80% of the questions takes 20% of the time that a fully comprehensive system would take; so it would be wise to develop a system that automated a client's first consultation, asking 80% of the necessary questions, and producing a report for the consultant. The consultant would then deal with the remaining 20% of the questions, saving time and hence money for either or both parties. There are some tasks for which 80% coverage would be nowhere near adequate, but if a company requires a system that saves time for their consultant while still keeping him or her in the loop, then 80% coverage is close to ideal.

# 5 Giving adequate replies to requests for information about automatic profiling

Automatic profiling is the derivation from personal data of information that may affect how or whether companies choose to deal with that person. The best-known use is for credit scoring: estimating the likelihood that if a person is granted credit, they will repay it.

The use of technology for automatic profiling has a long history. Financial institutions have used statistics-based credit scoring systems for many years. This is a field where machine learning techniques have been widely used, often achieving decision making accuracy well in excess of the rates achieved by human decision makers.

GDPR introduces numerous new restrictions on automatic profiling if it is used for decision making. Articles 13-15 require that the data subject be given a variety of information about their rights regarding the storage and processing of their personal data, including "the purposes of the processing for which the personal data are intended." If their personal data is being used for "automatic decision making including profiling" then they are also entitled to "meaningful information about the logic involved, as well as the significance and the envisaged consequences of such processing for the data subject." Article 22 spells out the purpose of these requirements: that "the data subject shall have the right not to be subject to a decision based solely on automated processing, including profiling, which produces legal effects concerning him or her or similarly significantly affects him or her."

So does GDPR require that data subjects are given a full explanation of how their profile score was calculated in order to take decisions about them? If a full explanation is needed, then machine learning techniques are effectively unusable, and a great deal of software will have to be re-implemented before May 2018.

There are different opinions on this question within the EU. However, the wording of the regulation is important here. It is significant that the GDPR separates the "meaningful information" and the "significance of processing". It is also significant that, while the GDPR does explicitly state that data subjects have a right to "obtain an explanation of the decision reached after such assessment," it does so in Recital 71 (which is non-binding) rather than in the legally binding Articles. What is mandatory according to GDPR is that data subjects can be given enough insight into the logic of a model and the significance of that logic to have the context necessary to intelligently opt out (Burt 2017).

What does this mean in practice for automatic profiling systems based on machine learning techniques? It means that such systems will require at least the following after 25 May 2018:

- A technical description of the model. Burt (2017) suggests describing where the data comes from; the method used; and how many features are selected for.

- An understanding of the decision(s) that the model is used to make and the consequences of a false positive or an omission. This will help to explain not only the logic of the decision but also its significance.

Both these explanations must be at a level suitable for understanding by a data subject who might want to opt out. Burt (2017) suggests that organisations might additionally want to provide a list of the benefits of automated processing versus the downsides of opting out to help data subjects make their decision.

Rule based AI systems can meet this requirement far more easily by providing an explanation of the rules that have been triggered and the consequences of each for the data subject's profiling score. However, there is a drawback to this approach: organisations may not want to release such a detailed explanation of their business logic. A competitor who was in possession of a large enough collection of such explanations would have all the knowledge they needed to reverse engineer the organisation's system.

The best way of meeting this requirement, therefore, may not be to introduce additional AI technology. Instead, organisations will need to write technical descriptions and decision consequence documents to describe how their existing technologies work.

# 6 Identifying and assessing a potential or actual data breach

Another area where AI technology might be of use to organisations during their data processing activities is in the identification and assessment of potential or actual breaches. Article 4 defines a "personal data breach" as "a breach of security leading to the accidental or unlawful destruction, loss, alteration, unauthorized disclosure of, or access to, personal data transmitted, stored or otherwise processed." This is a more-wide-ranging definition than even the equivalent laws in US states, most of which do not consider a data breach to be actionable unless the data is exposed in public or in the course of criminal activity.

Since there is a requirement to identify data breaches that have not been publicly exposed, it is the organisation's responsibility to recognise when a breach of security has occurred. This means that the monitoring of data security is a key task.

If a breach does occur, data controllers must notify the competent supervisory authority without undue delay (i.e. within 72 hours unless reasons can be supplied why this was not feasible). However, notice is not required if "the personal data breach is unlikely to result in a risk for the rights and freedoms of natural persons" (Article 33). So organisations may well find themselves needing to perform an assessment of whether notification is required in 72 hours or less with significant financial risks if their assessment produces an incorrect answer either way.

Monitoring and assessment are both tasks that can be supported by AI technology. There are numerous examples of both rule-based systems and machine learning systems that support both monitoring (e.g. Thompson et al 2015; Vyas et al 2012) and assessment (e.g. Kingston 1991, Dhurandhar et al 2015). In this case, the inputs to the monitoring activity consist of whatever data and information is collected by the organisation's security system; so an organisation with a sophisticated network monitoring system might consider a machine learning approach, while an organisation that relies on managers to monitor employees' compliance with policies and procedures would probably benefit more from a checklist approach similar to that described in Section 3. Perhaps the checklist could be implemented as a mobile app that managers can use while at employees' desks.

The assessment of whether a breach of security, resulting in a potential or actual loss of personal data, needs to be reported to the supervisory authority is a risk assessment task, and so the arguments for an underlying AI system are similar to those discussed in Section 4. The 72-hour time limit provides a strong incentive for organisations to invest in an AI system which will be always at their disposal, rather than having to engage legal or consultancy advice in a hurry. Since the exception clause in Article 33 focusses on the likely effect of the breach on data subjects, rather than on the company's own policies, the AI system will need minimal customisation to a company's particular circumstances; so it is likely that an AI software firm might develop a package that multiple organisations can purchase and use.

The question of legal liability arises here; if the system makes an incorrect recommendation, who is legally at fault? Kingston (2016) discusses various models under which an AI system might be considered legally liable. Since GDPR appears to be largely based on strict liability, and there seems no reason to consider an AI risk assessor as an 'innocent agent' that has been incited to illegal activity unknowingly (i.e. used in a way that it was not designed for), then there is a significant likelihood that an AI system that makes an incorrect recommendation would indeed be found liable for the breach of GDPR. Because of this, any software vendor selling an AI system to help assess whether a security breach needs to be reported would be strongly advised to follow the 80-20 model recommended in Section 4, and to make it clear to purchasers that the system is a decision support system, designed to save their legal experts time in their risk assessments rather than to automate every part of the decision making process. This should provide the software vendor with a legally valid reason to assign the final responsibility for the assessment decision to the human expert rather than to the AI software.

# 7 Conclusion

The GDPR raises significant business risks for organisations that carry out data processing. AI technology can help to mitigate those risks by supporting risk assessments, monitoring and compliance checking with best practice knowledge; with time savings in availability of advice; and in asking all and only the relevant questions.

Compliance-based systems require following well-documented rules and having explicit reasons for doing so, or for choosing not to do so. Also, the GDPR regularly requires explanations of reasoning, or data processing organisations may require explanations of an AI system's reasoning to convince them that the legal arguments it is making should stand up to scrutiny. For these reason, rule-based technology is more appropriate than machine learning technology in most. However, in the one aspect of GDPR that actually requires explanations of logic to be given to data subjects who request it, it may be undesirable to give full explanations for reasons of business confidentiality.

Some of the decisions that the GDPR requires are one-time decisions for a data processing organisation, which means that the only organisations likely to make a business case for an AI system to support those decisions are consultancies who advise multiple organisations on GDPR compliance. However, all organisations are required to monitor continually for security breaches and, if a potential or actual breach occurs, to assess swiftly whether they need to notify the relevant supervisory authority. Monitoring tasks may be better carried out by machine learning than by rule based technology, especially if there is a need to detect unforeseen events or unknown patterns of symptoms.